# Probability and Asset Updating using Bayesian Networks for Combinatorial Prediction Markets

**Wei Sun**
Center of Excellence
in C4I
George Mason University
Fairfax, VA 22030

**Robin Hanson**
Department of Economics
George Mason University
Fairfax, VA 22030

**Kathryn B. Laskey**
Department of Systems
Engineering and
Operations Research
George Mason University
Fairfax, VA 22030

**Charles Twardy**
Center of Excellence
in C4I
George Mason University
Fairfax, VA 22030

## Abstract

A market-maker-based prediction market lets forecasters aggregate information by editing a consensus probability distribution either directly or by trading securities that pay off contingent on an event of interest. Combinatorial prediction markets allow trading on any event that can be specified as a combination of a base set of events. However, explicitly representing the full joint distribution is infeasible for markets with more than a few base events. A factored representation such as a Bayesian network (BN) can achieve tractable computation for problems with many related variables. Standard BN inference algorithms, such as the junction tree algorithm, can be used to update a representation of the entire joint distribution given a change to any local conditional probability. However, in order to let traders reuse assets from prior trades while never allowing assets to become negative, a BN based prediction market also needs to update a representation of each user's assets and find the conditional state in which a user has minimum assets. Users also find it useful to see their expected assets given an edit outcome. We show how to generalize the junction tree algorithm to perform all these computations.

## 1 INTRODUCTION

Prediction is a fundamental task for AI systems. There is strong theoretical and empirical support for the superiority of ensemble forecasts over individual forecasts (Solomonoff, 1978). While weighted forecasts are theoretically optimal, it has been surprisingly difficult to beat a simple unweighted average. Prediction markets have emerged as a simple and robust way to give high-performing forecasters more influence on the aggregated forecast. Not only do prediction markets typically out-perform both individual human forecasters and naïve unweighted averages (c.f., Chen and Pennock 2010), they show promise as an information combination method for machine aggregation as well. Barbu and Lay (2011) show that combining machine learners with prediction markets often outperforms top ensemble predictors like Random Forest on UCI datasets. A key advantage is that "the market mechanism allows the aggregation of specialized classifiers that participate only on specific instances." That is, learners can opt to bid only on cases they understand. So while most prediction markets use only human traders, they need not. Indeed, smart traders use algorithms, and Nagar and Malone (2011) provide strong evidence that human-machine combinations outperform either: machines do well when the rules hold, while humans recognize "broken leg" situations where they don't.

We focus on market-scoring-rule systems which create prediction markets from the sequential application of proper scoring rules, sidestepping impossibility theorems that apply to simultaneous aggregation of forecasts (Hanson, 2003; Chen and Pennock, 2010). Independent of whether traders are human and/or machine, we address the question of how to make the market itself more expressive by allowing combinatorial trades. While traditional prediction markets ignore dependencies among events, combinatorial prediction markets explicitly consider and exploit dependencies among base events. As Chen and Pennock (2010) argue:

> Why do we need or want combinatorial-outcome markets? Simply put, they allow for the collection of more information. Combinatorial outcomes allow traders to assess the correlations among base objects, not just their independent likelihoods, for example the correlation between Democrats winning in Ohio and Pennsylvania.

We prove some new results regarding combinatorial prediction markets and report on an implementation using Bayesian networks.

## 1.1 PREDICTION MARKETS

Prediction markets make probabilistic forecasts by allowing participants to trade contingent assets. Prices in such a market can be interpreted as probabilities: if an asset paying $1 contingent on event $E$ is currently selling for $0.75, then the current market probability of $E$ is 75%. Prediction markets are an increasingly popular way to aggregate information and judgments from groups (Tziralis and Tatsiopoulos, 2007). Traders self-select to speak on the topics they think they know best. Those with more knowledge achieve greater influence by acquiring more assets, and market prices inform everyone of trader information.

In a market-maker-based prediction market, an automated trader stands ready to buy or sell assets on any relevant event. The prices it offers can be seen as a current trader consensus on the probabilities of those events, and trades can be seen as edits of consensus probabilities. In a logarithmic market scoring rule based (LMSR-based) prediction market, the market maker varies its price exponentially with the quantity of assets it sells. Tiny trades are fair bets at the consensus probabilities (Hanson, 2003). Larger trades change the consensus probabilities; we call such trades "edits." Users make edits in an attempt to maximize assets, thereby forming a consensus distribution that aggregates information from all market participants.

## 1.2 COMBINATORIAL PREDICTION MARKETS

In a combinatorial prediction market, one can trade on any event that can be specified as a combination (e.g. 'and' or 'or') of a base set of events. A market-maker-based combinatorial prediction market, therefore, declares a complete consistent probability distribution over a combinatorial space of events, and lets participants edit any part of that distribution. In a combinatorial LMSR-based market, users can make conditional bets that satisfy intuitive independence properties. For example, letting "$\neg$" denote "not", a trader who increases the value of $p(A|B)$ gains if $B$ and $A$ occur and loses if $B$ and $\neg A$ occur. Such an edit changes neither $p(B)$ nor $p(A|\neg B)$ and the trader neither gains nor loses if $\neg B$ occurs (Hanson, 2007).

With a large set of base events, the number of event combinations becomes astronomical, making it intractable in general to compute market prices and trades. In particular, it is in general NP-hard to maintain correct LMSR prices across an exponentially large outcome space (Chen et al., 2008a).

One way to achieve tractability is to limit the complexity of the consensus probability distribution by using a factored representation of the joint distribution. Examples include Bayesian networks (BNs) and Markov networks, which admit standard algorithms to efficiently compute conditional marginals and perform evidential updating (e.g. Pearl 1988; Jensen 1996; Shachter et al., 1990). Graphical models are widely used in many applications, often achieving tractable inference in networks with thousands of variables.

Bayesian networks have been used to represent joint distributions in prediction markets. Chen et al. (2008b) used a BN to represent prices in a tournament, and Pennock and Xia (2011) used a BN to represent probabilities in a LMSR-based combinatorial prediction market.

Pennock and Xia (2011) proved that probabilities can be updated in polynomial time for edits which do not violate the conditional independence assumptions of a decomposable BN of fixed treewidth. However, they do not show how to accomplish two other tasks that are important in practical prediction markets. They do not show how to let traders reuse assets purchased in previous trades to pay for new trades. They also do not show how to calculate a trader's expected assets to see if the trader is already 'long' or 'short' on an issue before making an edit.

## 1.3 REUSING ASSETS

In the LMSR framework of Pennock and Xia (2011), each edit takes the form of a participant paying cash to a market maker to obtain an event contingent asset, which pays cash if a certain event happens. A trader who has run out of cash is not permitted to make any more trades. Yet the assets one has obtained from prior trades are often sufficient to guarantee many more trades. For example, suppose a user buys an asset "Pays $10 if $A$," and then later buys an asset "Pays $10 if $\neg A$." Because one of these assets is guaranteed to pay off, the two assets are together worth $10 in cash, and could be used to support future purchases. However, in a naïve implementation, this $10 is unnecessarily tied up until the truth-value of $A$ is resolved. While we might imagine that a system could easily notice the guaranteed payoff and trade those assets in for cash, it would be difficult to notice more complex combinations of trades worth a guaranteed amount of cash.

In a market that allows conditional trades, a consensus probability $p(A|B) = x$ corresponds to a market

price of $$x$ for a trade that pays $1 if events $B$ and $A$ both occur, pays nothing if $B$ and $\neg A$ both occur, and is called off (returning the purchase price to the user) if $\neg B$ occurs. Although Pennock and Xia (2011) do not use conditional securities, conditional probabilities are established by trading securities that depend on joint states. A consensus probability $p(A|B) = x$ corresponds to a market price of $$x$ to purchase two separate securities, one paying $1 if $B$ and $A$ both occur and the other paying $$x$ if $\neg B$ occurs.

Now, suppose a user wants to trade on event $A$ given $N$ mutually exclusive conditions $B_i$, where the current market probabilities are $p(A|B_i) = x_i, i = 1 \ldots N$. Without asset reuse, such a user would have to purchase $N$ separate pairs of assets, where the $i^{th}$ pair costs $$x_i$ and pays $1 if $A$ and $B_i$ occur, 0 if $\neg A$ and $B_i$ occur, and $x_i$ if $\neg B_i$ occurs. The total purchase price of $\sum x_i$ would be tied up until one of the $B_i$ occurred, although the collection of assets is guaranteed to pay off at least $\sum \$x_i - \max\{\$x_i\}$. If $N$ is large and the probabilities are non-negligible, a considerable sum could be unnecessarily tied up.

## 1.4 OUR CONTRIBUTIONS

In order to be able to reuse assets, we first need to represent the user's assets in a form that allows efficient computation to find the minimum asset state after the user's edits. Further, a factored representation of assets will provide significant savings in space and improve efficiency of asset management. In this paper, we show how to exploit the junction tree to efficiently maintain a representation of a trader's state-dependent assets, i.e., for each state the final cash this trader would hold if this state were revealed in the end to be the actual state. We also show how to use these data structures to efficiently find the largest amount by which the user could raise or lower the probability of an event of interest, before the change might result in the trader holding negative assets in some state. Keeping edits within these limits ensures that traders can reuse assets while never 'going broke.' Asset reuse enables more efficient information aggregation (i.e., users can make more trades before running out of assets) for a given amount of assets.

A trader about to make an edit also usually finds it useful to know whether she is 'long' or 'short' on the issue she is about to trade. For example, a user editing the value of $p(A|B)$ might like to know whether she should currently expect to gain more if $B$ and $A$ happens, or if $B$ and $\neg A$ happens. Learning that she already stands to gain more if $A$ and $B$ happens should make a risk-averse trader more reluctant to raise the value of $p(A|B)$, thereby acquiring more such assets. In this paper we show how to efficiently maintain a representation of each trader's expected assets, where the expectation is with respect to the market consensus probabilities. We also show how to efficiently calculate the conditional expectations relevant for being 'long' or 'short' on a given edit.

The key word here is "efficiently". Existing combinatorial implementations follow the naïve joint-state enumeration described by Hanson (2007), and are therefore limited to at most about 20 related binary variables. Pennock and Xia (2011) proved that one can use Bayesian networks for combinatorial prediction markets. However, they treat an inverse system where assets are easy to calculate and prices are the bottleneck. Further, they did not have an implementation of their method. We use a single Bayesian (or Markov) network for representing the market probability distributions, stored as clique potentials in the associated junction tree. Further, we prove that assets can be represented using the same factorization as the joint distribution. Thus, our method maintains a parallel junction tree data structure for each user's assets. We describe algorithms for updating these asset junction trees when users make edits. Asset factorization is an important advance in space and computational efficiency. We have developed a complete MATLAB implementation and a partial Java implementation of our algorithms. To our knowledge, ours is the first published implementation of BN-based combinatorial prediction markets.

Although arbitrary graphical probability models are of course intractable, in this paper we show how to adapt the junction tree algorithm to perform the desired computations in models whose treewidth is not too large. It follows trivially that for a fixed treewidth, complexity is polynomial in the number of variables and linear in the number of simultaneous edits. This is a major improvement over existing naïve implementations, which are equivalent to a fully-connected graph and therefore exponential in the number of variables. We describe numerical experiments that demonstrate the anticipated exponential savings in space and time for given bounds on treewidth.

*Organization of this paper:* The next section states our definitions and notation. Section 3 briefly describes relevant work on Bayesian networks and the junction tree algorithm. Section 4.3 presents our probability and asset updating algorithm. In Section 5, we first walk through the algorithm using a simple 3-node BN model. We then examine the scalability of our algorithm in a simulation study. Sections 6 and 7 provide discussion and conclusion.

## 2  DEFINITIONS AND NOTATION

Capital letters such as $A, B_i, X$ denote random variables. Bold capital letters (e.g., $\mathbf{A}, \mathbf{B}_i, \mathbf{X}$) denote vectors of random variables. Corresponding lowercase letters (e.g., $a, b_i, x, \mathbf{a}, \mathbf{b}_i, \mathbf{x}$) denote particular instantiations of the random variables. Unless stated otherwise, symbols $p$ and $\phi$ denote probability distributions and likelihoods respectively. $\mathcal{B}$ represents a BN, $\mathcal{T}$ the corresponding clique tree, and $\mathbb{C}, \mathbb{S}$ the set of cliques and set of separators, respectively. The conditional probability distribution of $X$ given $Y$ is denoted by $p(X|Y)$. For a Bayesian network on random variables $\mathbf{X} = \langle X_1, X_2, \ldots, X_n \rangle$, we use $Pa(X_i)$ to denote the parents of $X_i$ in the directed acyclic graph corresponding to the BN. Thus, the conditional probability distribution (CPD) of the random variable $X_i$ is denoted by $p(X_i|Pa(X_i))$. We denote negation with "$\neg$", so $(X = \neg x) \equiv (X \neq x)$.

Each user $u$ has an asset value $S_\mathbf{x}$ associated with each joint outcome $\mathbf{x}$. When necessary to index assets by user, we add a superscript, $S_\mathbf{x}^u$. The expected assets $\bar{S}$ for user $u$ is the user's net worth:

$$\bar{S}^u = \sum_{\mathbf{x} \in \Omega} p(\mathbf{x}) S_\mathbf{x}^u \qquad (1)$$

where $\Omega$ is the Cartesian product of the state spaces of all random variables. We denote the cardinality of the joint state space by $L = |\Omega|$.

## 3  PRELIMINARIES

If we represent the prediction market's probability distribution as a Bayesian network, then edits in the prediction market correspond to soft evidence on variables in the Bayesian network (see Section 3.1 below), and probability updating can be done using standard algorithms. Allowable edits are determined by the user's assets, and in particular that state which has the fewest assets for the edit being considered. We show below that the same junction tree can be used to represent a factorization of both probabilities and assets. The minimum asset state can be found by using min-propagation in the asset junction tree. That can be done is a consequence of a theorem by Dawid (1992) showing that min-propagation works for any function of the probabilities in a junction tree. We use this fact in Section 4.3.

A Bayesian network $\mathcal{B}$ factors the joint distribution for the random vector $\mathbf{X}$ into a product of local distributions:

$$p(\mathbf{x}) = \prod_{1 \leq k \leq n} p(X_k = x_k | \mathbf{X}_{Pa(X_k)} = \mathbf{x}_{Pa(X_k)})$$

This factorization in turn can be compiled into an undirected tree structure called a junction tree. The junction tree is composed of cliques and separators, such that $p$ is the product of all clique marginal distributions divided by the product of all separator marginal distributions:

$$p(\mathbf{x}) = \frac{\prod_{c \in \mathbb{C}} p_c(\mathbf{x}_c)}{\prod_{s \in \mathbb{S}} p_s(\mathbf{x}_s)}. \qquad (2)$$

Here, $\mathbf{x}_c$ and $\mathbf{x}_s$ denote the states of the variables in clique $c$ and separator $s$, respectively, and $p_c$ and $p_s$ are the marginal distributions for the clique and separator variables, respectively. The junction tree algorithm (Lauritzen and Spiegelhalter, 1988) uses this transformation to perform exact inference on the BN. We note that although we focus on BNs, our algorithms apply equally well to any representation that can be compiled into a junction tree.

### 3.1  SOFT EVIDENCE & BELIEF UPDATING

General BN inference computes the posterior distribution given observations, also known as hard evidence. In prediction markets, we need to update market distributions given user edits that revise some non-extreme $p$ to another non-extreme $p'$ on variables. That is, evidence from user edits is in general uncertain. The literature considers two kinds of uncertain evidence. *Soft* evidence specifies a new probability distribution of the variable regardless of its previous distribution (Koski and Noble, 2009; Langevin and Valtorta, 2008; Valtorta et al., 2002), whereas *virtual* evidence, also known as *likelihood* evidence, represents the relative likelihood of the evidence given the true state. These likelihood values do not necessarily sum to 1 over all states. Virtual evidence is often implemented as observations on a hidden "dummy" node as the child node of the node on which virtual evidence is specified. In our case, we implement user edits as soft evidence. We use $\phi(X)$ to denote soft evidence on variable $X$. Usually, soft evidence on a single variable can be implemented as virtual evidence (Pearl, 1990), allowing standard junction tree inference algorithms to apply.

In our prediction market, a user edits the probability that the target variable $T$ is in one of its states $t$, for example, changing $p(T = t) = a$ into $p(T = t) = b$. We then assign $1 - b$ to $T$'s other states in proportion to their previous probabilities, and represent this edit as soft evidence on $T$. Similarly, if the user would like to edit $p(T = t|\mathbf{A} = \mathbf{a})$ conditional on the values of other variables $\mathbf{A} = \mathbf{a}$, we represent this conditional edit as conditional soft evidence. An unconditional edit corresponds to an empty assumption set $\mathbf{A} = \emptyset$.

## 3.2 MIN-CALIBRATION OF JUNCTION TREE

In the standard junction tree algorithm for discrete BNs, marginalization is done by summing out variables. When we replace summation with maximization in the propagation algorithm, a max-calibrated junction tree with max potentials for all cliques will be returned. It is then straightforward to find the configurations over all states with maximum joint probability based on these max-potentials. Similarly, min-calibration of the junction tree can be performed by replacing summation with minimization, and the minimum probability configuration can be found accordingly. Further, Dawid (1992) proved that min/max-calibrations are valid for functions on the potentials. We will use this fact to find the boundary conditions for a user's assets, thus allowing asset reuse while preventing the possibility of assets going negative.

## 4 PROBABILITY AND ASSET UPDATING ALGORITHM

Once again, any edit in the prediction market is implemented by asserting soft evidence in the corresponding BN. We use the junction tree inference algorithm to update the consensus joint probability distribution after each edit. Then we use LMSR as the market maker to update the user's assets accordingly. We assume a market trading on purely discrete variables and so the representing BN is a purely discrete network. After updating, each clique in the junction tree maintains the correct joint distribution of variables in the clique.

A given user has assets $S_\mathbf{x}$ associated with every joint state $\mathbf{x}$ of the domain variables. LMSR dictates that assets change in proportion to the log of the ratio of probabilities:

$$\Delta S_\mathbf{x} = b \ln \frac{p'(\mathbf{x})}{p(\mathbf{x})},$$

where $p'(\mathbf{x})$ is the new probability for the joint state $\mathbf{x}$ arising from the user's edit, $p(\mathbf{x})$ is the previous probability, and $b$ is a constant defining the unit of currency. Allowed changes must respect the rule that the minimum across all states of the user's assets must not be allowed to drop below zero, i.e., $\min_x S_x \geq 0$. After an edit to the consensus distribution, the asset data structure changes for the user making the edit; the asset data structures for all other users remain unchanged.

## 4.1 ASSET FACTORIZATION

It is convenient to define a transformation $q(\mathbf{x})$ of the assets $S_\mathbf{x}$, such that

$$S_\mathbf{x} = b \ln(q(\mathbf{x})). \tag{3}$$

We then have

$$\frac{q'(\mathbf{x})}{q(\mathbf{x})} = \frac{p'(\mathbf{x})}{p(\mathbf{x})}, \tag{4}$$

where $q(\mathbf{x})'$ is the updated asset for joint state $\mathbf{x}$, corresponding to the probability change $p'(\mathbf{x})$. The derivation is:

$$\begin{aligned}
S'_\mathbf{x} &= S_\mathbf{x} + \Delta S_\mathbf{x} \\
&= S_\mathbf{x} + b \ln \frac{p'(\mathbf{x})}{p(\mathbf{x})} \\
&= b \ln(q(\mathbf{x})) + b \ln \frac{p'(\mathbf{x})}{p(\mathbf{x})} \\
&= b \ln(q'(\mathbf{x}))
\end{aligned}$$

$$\Rightarrow b \ln(q'(\mathbf{x})) = b \ln(q(\mathbf{x})) + b \ln \frac{p'(\mathbf{x})}{p(\mathbf{x})}$$

$$\Rightarrow \ln(q'(\mathbf{x})) - \ln q(\mathbf{x}) = \ln \frac{p'(\mathbf{x})}{p(\mathbf{x})}.$$

Therefore, the identity $\ln a - \ln b = \ln a/b$ establishes Equation (4).

To interpret $b$, note that if the user changes the probability at state $\mathbf{x}$ from $p(\mathbf{x})$ to $p'(\mathbf{x}) = 1$ and $\mathbf{x}$ turns out to be true, the user gains $\Delta S_\mathbf{x} = -b \ln p(\mathbf{x})$. The maximum possible gain is therefore $-b \ln p(\mathbf{x}_*)$, where $\mathbf{x}_*$ is the minimum probability state. If all states start out equally likely, i.e., $p(\mathbf{x}) = 1/L$ for all $\mathbf{x}$, then the maximum gain to users, and the maximum loss to the market maker, is $b \ln L$, where $L$ is the total number of states. Therefore to bound losses to be no more than $M$, we usually initialize all market states to be equally likely at the start of trading and set $b = M/\ln L$.

Because $q$ starts out independent of the state and changes in proportion to changes in $p$, we can decompose $q$ in a similar manner to the decomposition of $p$, shown in Equation (2). Specifically,

$$q(\mathbf{x}) = \frac{\prod_{c \in \mathbb{C}} q_c(\mathbf{x}_c)}{\prod_{s \in \mathbb{S}} q_s(\mathbf{x}_s)}, \tag{5}$$

where $q_c$ and $q_s$ are local asset components defined on the clique and separator variables, respectively. Notice the similarity to (2). This factored representation for assets is preserved long as edits are confined to variables in the same clique, i.e., trades are *structure preserving* (Pennock and Xia, 2011). This allows us to do all calculations locally in every clique and separator, because Equation (4) is valid for the joint space of each clique and separator. Namely, we can establish the same junction tree structure for the assets $q$ and make local updates when edits occur. When there is a probability edit, we propagate the soft evidence in the junction tree to obtain the correct probability update for every clique and separator. For structure preserving trades, assets can be updated simply by choosing a clique $c$ containing the variables being traded and multiplying $q_c(\mathbf{x}_c)$ by the probability ratio:

$$q'_c(\mathbf{x}_c) = q_c(\mathbf{x}_c) \frac{p'_c(\mathbf{x}_c)}{p_c(\mathbf{x}_c)}. \tag{6}$$

Combining this with Equation (5) gives the same result as Equation (4). However, we usually do not need to compute global assets $q$ for each possible joint state. For space and computational efficiency, a representation of the user's assets is stored locally in cliques of the asset junction tree. This asset junction tree is used to compute the minimum asset value and its associated state, as well as the expected value user's expected assets.

To ensure that the user's assets remain non-negative in all states, we must place limits on the edits a user is allowed to make. Equivalently, no edit may allow the transformed assets $q$ as defined in Equation (3) to become less than 1. We must find the limits on edits beyond which the probability change will result in negative assets in some state. Let us assume that the user is editing $p(T = t | \mathbf{A} = \mathbf{a})$, denoted as $p^t$, to $p^\#$. Let $m_t$ denote her current minimum $q$ given $(\mathbf{A} = \mathbf{a}, T = t)$. Let $m_{\neg t}$ denote her current minimum $q$ given $(\mathbf{A} = \mathbf{a}, T \neq t)$. If $(\mathbf{A} = \mathbf{a}, T = t)$ occur after the edit, we have to ensure that the updated minimum $q^\#$ remains greater than 1. That is:

$$q^\# = m_t \frac{p^\#}{p^t} \geq 1.$$

Then

$$p^\# \geq \frac{p^t}{m_t}.$$

Similarly, if $(\mathbf{A} = \mathbf{a}, T \neq t)$ occurs after the edit, the updated minimum assets $q^u$ must remain greater than 1. That is:

$$q^u = m_{\neg t} \frac{1 - p^\#}{1 - p^t} \geq 1.$$

It follows that

$$p^\# \leq 1 - \frac{1 - p^t}{m_{\neg t}}.$$

Summarizing the above results, the allowable edit range for $p(T = t | \mathbf{A} = \mathbf{a})$ is:

$$\left[ \frac{p(T = t | \mathbf{A} = \mathbf{a})}{m_t}, 1 - \frac{1 - p(T = t | \mathbf{A} = \mathbf{a})}{m_{\neg t}} \right]. \tag{7}$$

Note that the minimum assets can be found by min-calibration over the asset junction tree, as mentioned in Section (3.2).

For every user, we maintain a separate asset junction tree in which the affected clique is updated only after this particular user makes an edit. Edits made by a given user will have no effect other users' assets. But because every edit changes the market probability, we update market distribution after each edit accordingly (where the updates are stored as clique potentials of the junction tree).

### 4.2 EXPECTED VALUE / SCORE

A user typically wants to know the expected value of her assets given the current market consensus prices. If she is contemplating an edit to event $A$, she would want to know what her expected assets will be if $A$ happens and if $\neg A$ happens. These expectations can be calculated efficiently given the factorization represented in the junction tree. The expected score is obtained as follows (recall that $c$ indexes cliques and $s$ indexes separators).

$$\bar{S} = \sum_c \sum_{x_c} S_c(\mathbf{x}_c) p_c(\mathbf{x}_c) - \sum_s \sum_{x_s} S_s(\mathbf{x}_s) p_s(\mathbf{x}_s), \tag{8}$$

where

$$S_c(\mathbf{x}_c) = b \ln(q_c(\mathbf{x}_c)), \tag{9}$$

and

$$S_s(\mathbf{x}_s) = b \ln(q_s(\mathbf{x}_s)). \tag{10}$$

This result is derived as follows. First, we substitute (5) into (3) and then substitute (9) and (10) into the result to obtain the following expression for $S_\mathbf{x}$:

$$S_\mathbf{x} = \left[ \sum_c S_c(\mathbf{x}_c) - \sum_s S_s(\mathbf{x}_s) \right]. \tag{11}$$

Next, we substitute (11) into (1), suppressing the superscript $u$, to obtain:

$$\begin{aligned} \bar{S} &= \sum_x p(\mathbf{x}) \left[ \sum_c S_c(\mathbf{x}_c) - \sum_s S_s(\mathbf{x}_s) \right] \\ &= \sum_c \sum_x S_c(\mathbf{x}_c) p(\mathbf{x}) - \sum_s \sum_x S_s(\mathbf{x}_s) p(\mathbf{x}). \end{aligned}$$

Letting $\mathbf{x}_{\neg c}$ and $\mathbf{x}_{\neg s}$ denote the states of the variables not in clique $c$ and separator $s$, respectively, we obtain:

$$\begin{aligned} \bar{S} &= \sum_c \sum_{\mathbf{x}_c} S_c(\mathbf{x}_c) \sum_{\mathbf{x}_{\neg c}} p(\mathbf{x}) - \sum_s \sum_{\mathbf{x}_s} S_s(\mathbf{x}_s) \sum_{\mathbf{x}_{\neg s}} p(\mathbf{x}) \\ &= \sum_c \sum_{x_c} S_c(\mathbf{x}_c) p_c(\mathbf{x}_c) - \sum_s \sum_{x_s} S_s(\mathbf{x}_s) p_s(\mathbf{x}_s), \end{aligned}$$

which establishes (8). Further, let

$$\bar{S}_c = \sum_{x_c} S_c(\mathbf{x}_c) p_c(\mathbf{x}_c),$$

and

$$\bar{S}_s = \sum_{x_s} S_s(\mathbf{x}_s) p_s(\mathbf{x}_s).$$

Then for brevity, we can write

$$\bar{S} = \sum_c \bar{S}_c - \sum_s \bar{S}_s.$$

Clearly, the values $\bar{S}_c$ and $\bar{S}_s$ represent local expected scores for cliques and separators, respectively. This local score decomposition demonstrates once again the beauty of factorization.

### 4.3 PROBABILITY AND ASSET UPDATING ALGORITHM

To summarize our procedure for updating probabilities and assets, please see Algorithm 1. The algorithm maintains a consensus probability distribution and an asset structure for each user. The probability distribution and the asset structures factor according to the same junction tree. After each edit, the assets of the user making the edit and the consensus distribution are updated.

---

**Algorithm 1** Update probability and user assets for a BN representing a combinatorial prediction market

**Require:** a BN model $\mathcal{B}$ over a set of domain variables $\mathbf{X}$ that represents the combinatorial prediction market joint distribution, the clique tree $\mathcal{T}$ corresponding to $\mathcal{B}$, consisting of cliques $\mathbb{C}$ and separators $\mathbb{S}$.
**Require:** The current market probability distribution $p$ represented by a probability junction tree.
**Require:** The current assets $q$ for user $u$ represented by an assets junction tree.
    **for** each conditional edit $p'$ on the target variable $T = t$ with assumptions $\mathbf{A} = \mathbf{a}$ by user $u$: **do**
      - Tell the user the expected scores $\bar{S}(T = t, \mathbf{A} = \mathbf{a})$ and $\bar{S}(T \neq t, \mathbf{A} = \mathbf{a})$ indicating the user's long/short status.
      - Calculate the edit limits for $p(T = t|\mathbf{A} = \mathbf{a})$ using Equation (7).
      - Allow user to trade $p'(T = t|\mathbf{A} = \mathbf{a})$ within the edit limits. And apply $p'(T = t|\mathbf{A} = \mathbf{a})$ as soft evidence to the junction tree.
      - Update probability distributions of cliques and separators to be $p'_c$, $p'_s$, $c \in \mathbb{C}, s \in \mathbb{S}$ by calling the junction tree inference algorithm.
      - Find the clique $c$ containing target variable $T$ and assumed variables $\mathbf{A}$, and update the assets clique corresponding to $c$ using Equation (6).
    **end for**
    **return** User's expected assets after the edit; user's min-$q$ value and its associated min-$q$ states.

---

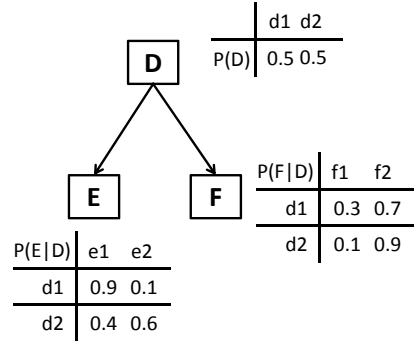

Figure 1: *BN-DEF*: An Example network With Three Binary Nodes $D, E,$ and $F$.

## 5 NUMERICAL EVALUATION

### 5.1 TEST CASES FOR *BN-DEF*

We developed a complete MATLAB implementation for our algorithm using BNT (Murphy, 2001) and currently we are developing a Java implementation in UnBBayes (Matsumoto et al., 2011). Our implementation requires the target variable and all assumed (conditioning) variables to belong to the same clique. Cross-clique conditioning is possible, but requires approximations (see Section 6).

In this section, we illustrate how the algorithm works using the simple 3-node BN model *BN-DEF*, shown in Figure 1. The CPDs for nodes in *BN-DEF* are shown in the picture next to the corresponding nodes. This network has only two cliques $\{D, E\}$ and $\{D, F\}$. We work through the algorithm, showing intermediate results.

We begin by initializing $q$ to be 100 for every cell in the asset tables. The scale parameter $b$ is specified as $10/\ln(100)$, making the initial asset score 10 for every state for all users. We imagine two users, Joe and Amy, making successive edits.

**Joe's first edit** - Suppose Joe thinks the current market probability of $E = e1$ (0.65) is not reasonable, and wants to increase it. Since this will be his first trade, his initial uniform $q$ makes his current position neutral because both min-$q$ values given $E = e1$ and $E \neq e1$ are equal to 100. Using Equation (7), we calculate his edit limits on $E = e1$ to be $[0.0065, 0.9965]$. Based on his private information, Joe chooses 0.8 as the new $p(E = e1)$. The market distribution is then updated such that marginal probabilities of $D, E, F$ are now $[0.58, 0.42], [0.8, 0.2]$, and $[0.22, 0.78]$ respectively. Joe's expected assets are $\bar{S} = 10.12$, and his min-$q$ is 57.14 at two min-states - $\{d2, e2, f1\}$, and $\{d2, e2, f2\}$.

**Amy's first edit** - Now Amy is interested in changing $p(D = d1|F = f2)$, which due to Joe's edit is currently 0.52. Again, since this will be Amy's first trade, she has neutral positions. Using Equation (7), we find her edit limits on $D = d1$ given $F = f2$ to be $[0.0052, 0.9952]$. Suppose Amy wants to move the probability to 0.7. Now, the market distribution is updated such that the marginal probabilities of $D, E, F$ are now $[0.72, 0.28], [0.85, 0.15]$, and $[0.22, 0.78]$ respectively. Note that $F$ is unaffected – it was assumed. Amy's expected assets are $\bar{S} = 10.11$, and her min-$q$ is 62.54 at two min-states - $\{d2, e1, f2\}$, and $\{d2, e2, f2\}$. Note that Amy's edit does not affect any other users' asset data.

**Joe's second edit** - Let us now consider an extreme edit example. Joe comes back and wants to see what will happen if he makes a big move on $p(E = e1|D = d2)$ (currently equal to 0.59). First of all, Joe finds that he has a long position for this trade since his $\bar{S}(E = e1, D = d2) = 10.45$, and $\bar{S}(E \neq e1, D = d2) = 8.79$. Second, his edit limits returned by the algorithm are $[0.0048, 0.9928]$. If he decides to move $p(E = e1|D = d2)$ to 0.99, then the marginal probabilities of $D, E, F$ can be updated to $[0.72, 0.28], [0.96, 0.04]$ and $[0.22, 0.78]$, respectively. Joe's expected assets after this trade are $\bar{S} = 10.67$, and his min-$q$ is 1.39 at two min-states - $\{d2, e2, f1\}$, and $\{d2, e2, f2\}$. Note that in this case, Joe's min-$q$ is very close to the threshold.

## 5.2 SCALABILITY STUDY

To investigate scalability, we first tested our algorithms on randomly generated networks of varying sizes. Our random networks were generated using BN-Generator 0.3 (Ide et al., 2004). We varied the number of variables from 30 to 960 and the treewidth bound from 5 to 20. For each combination of number of variables and the treewidth, three random BNs were generated. All random variables were binary. For each randomly generated network, we calculated the system lock time, defined as the CPU time required to update the probability distribution after an edit is confirmed by the user. While the system is locked, new edits have to be rejected or queued; thus lock time is a key performance metric. We ran this test on our Java implementation. Although not fully complete, the Java implementation performs probability updating, the part of the calculation required to determine the system lock time.

Figure 2 shows the results of our scalability study. As is well known, the complexity of the junction tree algorithm – and therefore the maximum lock time – is polynomial in the treewidth. We can see from the figure that the system can easily handle binary networks of treewidth 10, but performance begins to degrade noticeably with treewidth 20. The algorithm is not very sensitive to increases in the number of variables when the treewidth is held constant.

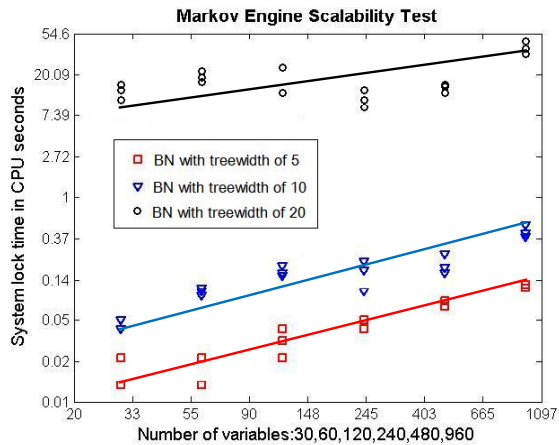

Figure 2: Edit Lock Time versus Network Size

We also investigated another key performance parameter, the potential rejection rate under different market environments, modeled by frequency of edits. As noted above, edits attempted while the system is locked may have to be rejected. For this study, we used ALARM (Beinlich et al., 1989), a 37-node BN with treewidth of 4 often used as a benchmark for graphical model algorithms. We chose ALARM because its size and treewidth are at the upper end of the range we expect for our live combinatorial prediction market. The objective of this test is to investigate how our approach performs with frequent edits by multiple users.

We simulated a market with 100 participants, and expected edits of 2/minute, 8/minute, and 30/minute, assuming a Poisson distribution for edit arrivals. For comparison, on Super Tuesday, InTrade had 18,629 trades from 780 unique users, for an average of 13 trades per minute. Prediction market trades are quite lumpy in both questions and time; we think our figures span a plausible range. The system lock time for ALARM is 0.3 seconds in our MATLAB implementation. We did not implement enhancements such as lazy propagation (Madsen and Jensen, 1998) that would give further edit-dependent reductions in lock time. 1000 edits were simulated by randomly chosen users from 100 market participants, given randomly chosen assumed variables and betting on randomly chosen target variable from randomly chosen cliques among the 27 cliques of ALARM's junction tree. If the randomly chosen clique had more than 3 variables, we then randomly chose two variables with their randomly chosen states to be assumed values on which the edit was conditioned. Inter-arrival times between edits were modeled by exponentially distributed variables with means of 2, 7.5 and 30 seconds respectively, representing market intensities of 30, 8 and 2 edits per minute, correspondingly. Table 1 presents the average number of rejects and the rejection rates for our experiments.

Table 1: ALARM: Simulation Results for 1000 Edits

| Market Intensity | Average rejects | Average rejection rate |
|---|---|---|
| 2 edits/minute | 11.3 | 1.2% |
| 8 edits/minute | 39.2 | 4% |
| 30 edits/minute | 142.6 | 14.3% |

In this network, the system could sustain an arrival rate of 8 edits/minute (11,520/day) with less than a 5% rejection rate. By queuing edits and rejecting only those whose terms have become worse for the user, this can be at least halved. A compiled language would be much faster for all the non-matrix operations, and with a good choice of library, not much slower at matrix operations.

## 6 NON STRUCTURE-PRESERVING EDITS

When we have conditional edits such that the target variable and the assumed variables are not in the same clique, then the edits are not structure preserving. In that case, if we keep our sparser structure, then the minimum KL-distance approximation to the true posterior distribution will be the one that has the same marginal distributions for all the cliques (Darwiche, 2009).

## 7 CONCLUSION AND FUTURE WORK

In this paper, we describe a method to update both the probabilities and assets in a combinatorial prediction market. The approach uses a junction tree compiled from the associated Bayesian network to represent the consensus probability distribution, and a structurally identical junction tree to represent each participant's assets. Allowable edits are calculated using min-propagation in the assets junction tree, and an allowable edit is chosen. The probability distribution is updated by applying soft evidence to the (shared) probability junction tree, and propagating. Assets are updated by simple multiplication in a single clique, without requiring a separate propagation. We developed a complete implementation of our algorithm and demonstrated its scalability by performing computational experiments on randomly generated BNs with varying sizes from 30 to 960 variables, and treewidth of 5 to 20. Further, we conducted a simulation study to illustrate the robustness of our method under varying market environments. Test results show that the system performs well for networks with treewidth of up to 10, and can sustain a simulated market with expected arrival rate of 8 edits/minute with less than 5% rejection rate.

Future work will measure the computational savings (or increase in network size) compared to a more straightforward combinatorial prediction market algorithm. We also plan use this technique in an online forecasting system, and to examine the tradeoff between a minimum-KL approximation (which may confuse users) and a less accurate but simpler set of assumptions.

**Acknowledgements**

The authors gratefully acknowledge support from the Intelligence Advanced Research Projects Activity (IARPA) via Department of Interior National Business Center contract number D11PC20062. The U.S. Government is authorized to reproduce and distribute reprints for Governmental purposes notwithstanding any copyright annotation thereon. Disclaimer: The views and conclusions contained herein are those of the authors and should not be interpreted as necessarily representing the official policies or endorsements, either expressed or implied, of IARPA, DoI/NBC, or the U.S. Government.

We also gratefully acknowledge the hard work of graduate student Shou Matsumoto for coding the Java implementation.